\documentclass[runningheads]{llncs}
\usepackage[T1]{fontenc}
\usepackage{graphicx} 
\usepackage[table]{xcolor}
\usepackage{array}        
\usepackage{xtab}
\usepackage{multirow}
\usepackage{hyperref}
\usepackage{subcaption}
\usepackage{graphicx}

\date{\today}

\author{Alex Interrante-Grant \and Andy Davis \and Heather Preslier \and Tim Leek}

\institute{MIT Lincoln Laboratory, Lexington MA, USA}

\authorrunning{A. Interrante-Grant et al.}

\title{On Training a Neural Network to Explain Binaries}

\begin{document}

\maketitle

\begin{abstract}
	In this work, we begin to investigate the possibility of training a deep neural
network on the task of binary code understanding. Specifically, the network
would take, as input, features derived directly from binaries and output
English descriptions of functionality to aid a reverse engineer in
investigating the capabilities of a piece of closed-source software, be it
malicious or benign. Given recent success in applying large language models
(generative AI) to the task of source code summarization, this seems a
promising direction.  However, in our initial survey of the available datasets,
we found nothing of sufficiently high quality and volume to train these complex
models. Instead, we build our own dataset derived from a capture of Stack
Overflow containing 1.1M entries. A major result of our work is a novel dataset
evaluation method using the correlation between two distances on sample pairs:
one distance in the embedding space of inputs and the other in the embedding
space of outputs. Intuitively, if two samples have inputs close in the input
embedding space, their outputs should also be close in the output embedding
space. We found this Embedding Distance Correlation (EDC) test to be highly
diagnostic, indicating that our collected dataset and several existing
open-source datasets are of low quality as the distances are not well
correlated. We proceed to explore the general applicability of EDC, applying it
to a number of qualitatively known good datasets and a number of synthetically
known bad ones and found it to be a reliable indicator of dataset
value\footnote{DISTRIBUTION STATEMENT A. Approved for public release. Distribution is unlimited.

This material is based upon work supported by the Under Secretary of Defense
for Research and Engineering under Air Force Contract No. FA8702-15-D-0001. Any
opinions, findings, conclusions or recommendations expressed in this material
are those of the author(s) and do not necessarily reflect the views of the
Under Secretary of Defense for Research and Engineering.

© 2024 Massachusetts Institute of Technology.

Delivered to the U.S. Government with Unlimited Rights, as defined in DFARS
Part 252.227-7013 or 7014 (Feb 2014). Notwithstanding any copyright notice,
U.S. Government rights in this work are defined by DFARS 252.227-7013 or DFARS
252.227-7014 as detailed above. Use of this work other than as specifically
authorized by the U.S. Government may violate any copyrights that exist in this
work.
}.
    
\end{abstract}

\section{Motivation}\label{sec:Motivation}
Binary programs are big and opaque and we typically know little, with
certainty, about their function except at the highest level of abstraction.
This fact is at variance with the obvious need to be able to state that this
code does this or that and no more. Malware is an extreme example of this need
and explaining what it does is currently the province of reverse engineers who
are expensive and in short supply. Automated means for accelerating or
assisting this activity would be of tremendous value.

In this work, we hoped to leverage recent advances in the use of Large Language
Models (LLMs) in processing, analyzing, and generating program binaries and
English prose~\cite{coda}\cite{neutron}. We imagined a tool that reverse
engineers could use to generate English descriptions of the functionality they
are attempting to reverse engineer. A user would provide input in the form of
some machine code and the tool would generate an English description such as
\emph{This function takes an array of 32 bit integers and returns the average
of the positive elements} as output.

We assert that the steps in building such a tool are as follows.

\begin{enumerate}
\item Find or assemble a train/test dataset.
\item Evaluate the quality of that dataset.
\item \label{step3} If the dataset is not good enough there are two avenues. Refine and repair if possible. Else, reject and start again.
\item If the dataset is judged acceptable, design and train a model.
\item Evaluate the model dataset via standard methods, i.e., cross-validation.
\end{enumerate}

Our efforts got as far as \ref{step3}. That is, we determined that the
datasets we had assembled and would have liked to proceed to use for training
and testing simply did not represent the concept required: consistent and
useful English descriptions of binary code. We believe that the ideas and
methodology we developed to allow us to confidently make such a claim are
independently of value and worthy of immediate reporting to the community at
this time. This is the chief result and business of this extended abstract.

\section{Existing Datasets}\label{sec:ExistingDatasets}
\label{sec:existing-datasets}

Machine learning models are, at best, only as good as the dataset(s) that they
are trained on. Here are what we consider to be the requirements for a dataset
from which to learn a model that takes binaries as input and outputs
explanations of functionality. 

\begin{enumerate}
\item \textbf{Code + prose explanations}. 
Code must be paired with explanations that are at a useful semantic level. For
example, overly high-level descriptions of functionality like \emph{This
function summons the final boss} are not as useful in a reverse engineering
context as lower-level descriptions that describe the business logic of code in
detail.
\item \textbf{Compilable source code}. 
This is not a hard requirement as binaries, in principle, would work fine.
However, source has two major advantages. First, we can spot check if
explanations correspond to code when we can easily read it. Second, source
means we can generate binaries and even diversify them by changing compilers,
flags, and using program transformations. 
\item \textbf{Many samples}. 
It is difficult to say with certainty but 100s or even thousands of examples
will not be enough to represent a concept of this sort: an english explanation
for this binary code. 
\end{enumerate}

\begin{table}[h!]
\caption{Existing Datasets for binary to explanation task learning}
\label{table:existing-datasets}
\begin{tabular}{|l|l|l|} 
\hline
Dataset Name                                      & Notes                                         & Viable? \\ \hline \hline
CodeXGLUE/Code-Text\cite{husain2019codesearchnet} & Mostly interpreted languages                  & No      \\ \hline
Deepcom-Java (BASTS)\cite{basts}                  & Java and Python only                          & No      \\ \hline
SPoC\cite{spoc2019}                               & Over 18K samples but explanations are         & No      \\
                                                  & ``pseudocode'' at wrong semantic level        &         \\ \hline
CONCODE\cite{CONCODE}                             & Java only                                     & No      \\ \hline
GitHub-Code                                       & C and C++ code but no explanations            & No      \\ 
                                                  & \scriptsize \texttt{https://huggingface.co/datasets/codeparrot/github-code} \normalsize & \\ \hline
CoNaLa\cite{yin2018mining}                        & Generated from StackOverflow posts but Python & No      \\ \hline
XLCoST\cite{zhu2022xlcost}                        & Problem summaries paired with implementations & Yes     \\
                                                  & 123K unique programs.                         &         \\ \hline 
HumanEval-X                                       & High quality explanations paired with         & No      \\
                                                  & implementations but only 820 samples          &          \\ 
	                                              & \scriptsize \texttt{https://huggingface.co/datasets/THUDM/humaneval-x} \normalsize & \\ \hline
Google Code Jam                                   & Problem statement ``explanations'' and        & No       \\
                                                  & submitted implementations. Many samples       &          \\
                                                  & but explanations are too high level           &          \\ 
                                                  & \scriptsize \texttt{https://codingcompetitions.withgoogle.com/codejam} \normalsize & \\ \hline \hline
\end{tabular}
\end{table}

We surveyed all datasets we could find, gleaned from a literature review done
by the authors~\footnote{Citations are provided for datasets with accompanying
papers; else, a URL for the dataset is listed in the table}.  The result of
that survey appears, in capsule form in Table~\ref{table:existing-datasets}.
From our review of the available data, we identified only \emph{one} possibly
viable candidate which met all three criteria: XLCoST.

\section{Constructing a Dataset}\label{sec:Dataset}
\label{sec:constructed-dataset}

While we might have proceeded directly with XLCoST as a dataset, we chose to investigate constructing a dataset inspired by CoNaLa, which used the popular programming question / answer website Stack Overflow as a source of annotated code\footnote{\texttt{https://stackoverflow.com}}.
XLCoST is a fairly small dataset and and it seemed likely that we could extract a much larger dataset from Stack Overflow. 

We began with an offline snapshot of Stack Overflow provided by the Kiwix project, which provides snapshots of popular websites\footnote{\texttt{https://www.kiwix.org}}. 
Our first step was to parse all pages tagged as either C or C++ and convert them to a format more conducive to automated processing. 
This process provided us with 1,107,347 pages.
Users provide questions and answers to Stack Overflow as free-form HTML without any semantic structure, with code intermixed with data and explanations.   
To make it easier for later stages of our processing to handle this unstructured data, we process raw pages into \emph{snippets} that contain the text contained within a single set of HTML tags indicating a section. 
A page consists of one or more snippets.

After we've parsed the questions and their answers into snippets, we attempt to compile binaries from the source code collected.
We used an algorithm inspired by CoNaLa, but made modifications for compilation to generate our candidate source files.
A single Stack Overflow page contains one question and one or more answers.
For the question and each answer, separately, we took all of the snippets that had been formatted as code and compute every permutation of contiguous concatenation. 
As an example, given the code snippets for a question or answer $a, b, c$, 
we would consider each of the following sequences of snippets as possibly compilable programs, or \emph{candidates}: 
$a, ab, abc, b, bc, c$.
We run each of these candidates through a preprocessor that attempts to fix or detect common errors.
We then inject each of the candidates into a few C and C++ templates that supply boilerplate code that is often left out when discussing code on Stack Overflow. 
Next, we run each candidate through a post processor to fix easy issues and discard obviously broken code.
Candidates are validated by ensuring that they compile and that they generate at least some code, as compared with a vacuuous program consisting only of \verb+int main(){}+.
If a single question or answer from page results in multiple valid candidates that compile, we retain the longest one. 
We generate one entry in our dataset per question that generates a valid binary and one per answer that generates a valid binary.
This means that, for a given Stack Overflow page, if it consists of a question and $N$ answers, all of which have at least one valid source candidate, then $N+1$ entries will be added to our dataset for that single page.

We also compose an English language explanation paired with each validated code candidate, consisting of a concatenation of all the non-code snippets in the question or answer from which the candidate was composed.

After the above processing, we're left with samples consisting of the binary code text section and associated English language explanation, i.e., samples that could be used to train or fine-tune a model. 
However, it is worth noting that our samples include some additional information, with associated Stack Overflow page metadata including upvotes, tags, and other features.
The final dataset consists of 73,209 samples.

\section{Evaluation Methodology}\label{sec:EvaluationMethodology}
\subsection{Embedding Distance Correlation (EDC) Method}
\label{sec:subEDCMethod}

\begin{figure}
\begin{centering}
\includegraphics[width=4in]{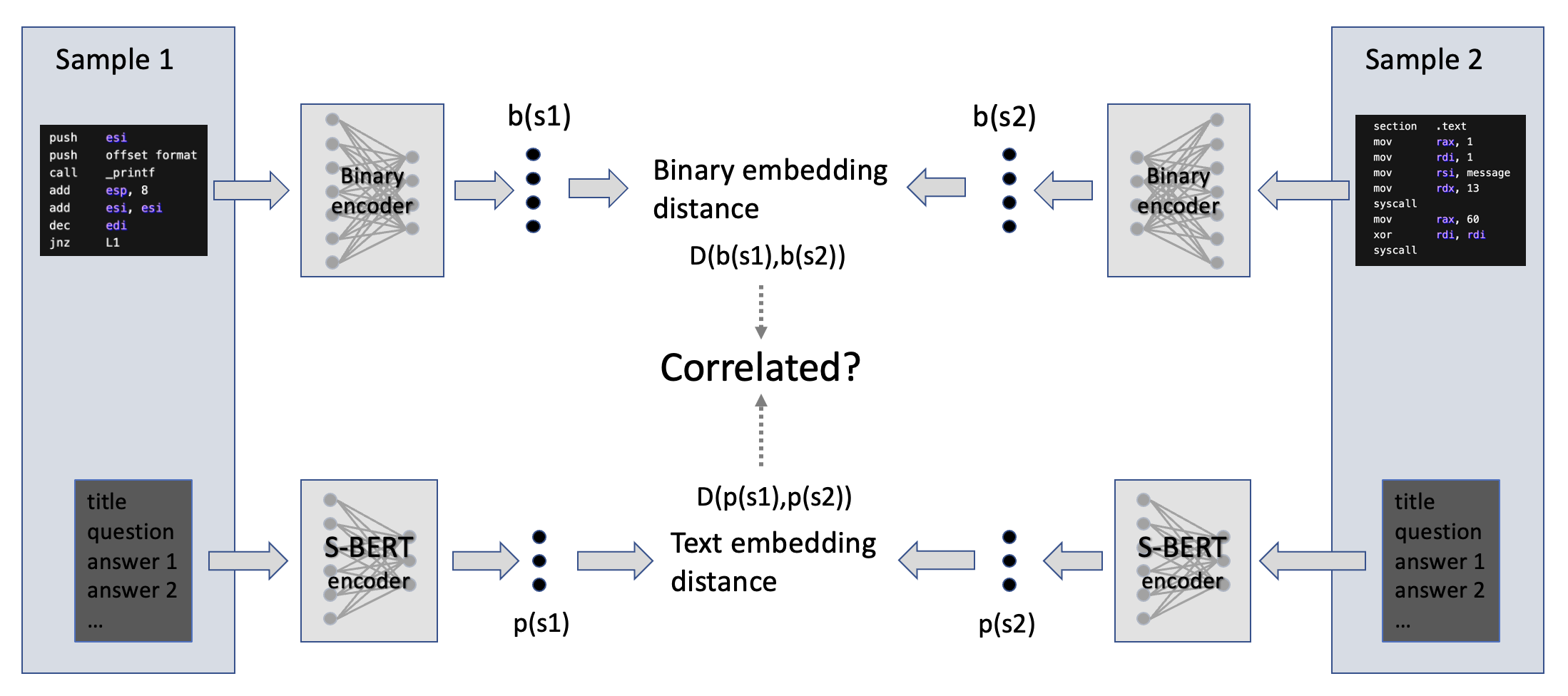}
\caption{For pairs of samples, the distance in the binary code 
embedding space and in the prose embedding space should be correlated if the
concept is learnable.}
\label{fig:binary-prose-dist-corr}
\end{centering}
\end{figure}

In order to evaluate the quality of a dataset, we designed a
novel methodology which measures the suitability of a
dataset for learned modeling independent of any model details, 
save that it be a transformer 
employing pre-trained embeddings. Our method, which we refer to as Embedding
Distance Correlation (EDC), considers randomly selected pairs of samples in a
dataset, generates embeddings using existing embedding
models in both the input and label domains, calculates the distances between the embeddings for each pair, and
finally computes the correlation of these distances over a large number of
pairs of samples. The algorithm for applying EDC to a dataset like Stack Overflow described in Section~\ref{sec:constructed-dataset} is as follows and is also depicted in Figure~\ref{fig:binary-prose-dist-corr}.

\small
\begin{enumerate}
\item Choose two samples from the dataset, $s_1$ and $s_2$
\item Generate the binary embedding vectors for both samples: $b(s_1)$ and $b(s_2)$
\item Generate the prose embedding vectors for both samples: $p(s_1)$ and $p(s_2)$
\item Compute the distance between the two binary embedding vectors $D(b(s_1), b(s_2))$
\item Compute the distance between the two prose embedding vectors $D(p(s_1), p(s_2))$
\item Repeat until a large number of pairs $D(b(s_1), b(s_2)), D(p(s_1), p(s_2))$ has been generated for statistical significance ($p < 0.05$) in the correlation of these distances.
\end{enumerate}
\normalsize

If the pairwise distances show a high degree of correlation 
then we can be certain that the concept represented by the dataset is
learnable, whereas a low correlation indicates either the dataset is of
poor quality or the embedding are low quality or not well matched to 
the input or labels. We can  mitigate against this latter issue
by using the best-performing, state-of-the-art embedding models 
in the relevant domains (binary code, and natural language prose 
in our case).

Our dataset evaluation method aligns well with the fundamental technology behind 
transformers, which consist of an encoder, which maps the model input domain into 
some embedding space, and a decoder, which maps another embedding space into the model 
output label range. EDC evaluates how effectively both the input features and the 
output labels can be embedded in their respective domains and how often samples 
which are close in the input space are also close in their output labels.

\subsection{Choice of Embeddings}
\label{sec:subChoiceOfEmbeddings}

For our application of EDC to the Stack Overflow and other datasets, we used the Sentence 
Transformers (SBERT) text embedding model~\cite{sentencebert} to embed code 
summaries, and the Instruction Trace Embedding Model 
(ITEM)~\footnote{This is a transformer-based, contrastively trained model that 
we developed internally and is not yet published. It is similar in concept and 
performance to BinShot~\cite{binshot}.} to embed the disassembled 
\texttt{.text} section of the sample binaries. Only the \texttt{.text} section of
the binary is considered so that binary headers, data, and metadata do not
affect the embedding.

\subsection{Human Expert Sanity Check}
\label{sec:subHumanExpertMethod}

To further validate our EDC method, we performed a manual survey of a
subset of the Stack Overflow dataset. We chose a random sample of 100 pairs of samples where
the binary code embedding distance was less than 0.2 and 100 additional samples
where the text description prose embedding distance was less than 0.5.
~\footnote{These thresholds were chosen by inspecting histograms of distances 
for pairs for each embedding. We observed a limited dynamic range of distance 
values for SBERT, for which a distance of 0.5, for this data, was considered 
quite close.}
These 100 pairs were divided up into four sets and presented to the four
authors of this report for evaluation. Adjudicators were asked to label each
pair in one of the following ways:

\begin{itemize}
\item \textbf{Agree}: Agree with distance computed; this pair is similar.
\item \textbf{Unsure}: Unsure if this distance make sense.
\item \textbf{Disgree}: Disagree with distance; pair seems unrelated.
\end{itemize}

This evaluation gives us some indication of if we, the authors, generally agreed
with the dataset labels and the embedding models in the EDC evaluation.

\section{Results}\label{sec:Results}
\subsubsection{Methodology Validation}
\label{sec:methodologyValidation}

\begin{figure}
\centering
\begin{subfigure}{0.4\textwidth}
  \includegraphics[width=2.5in]{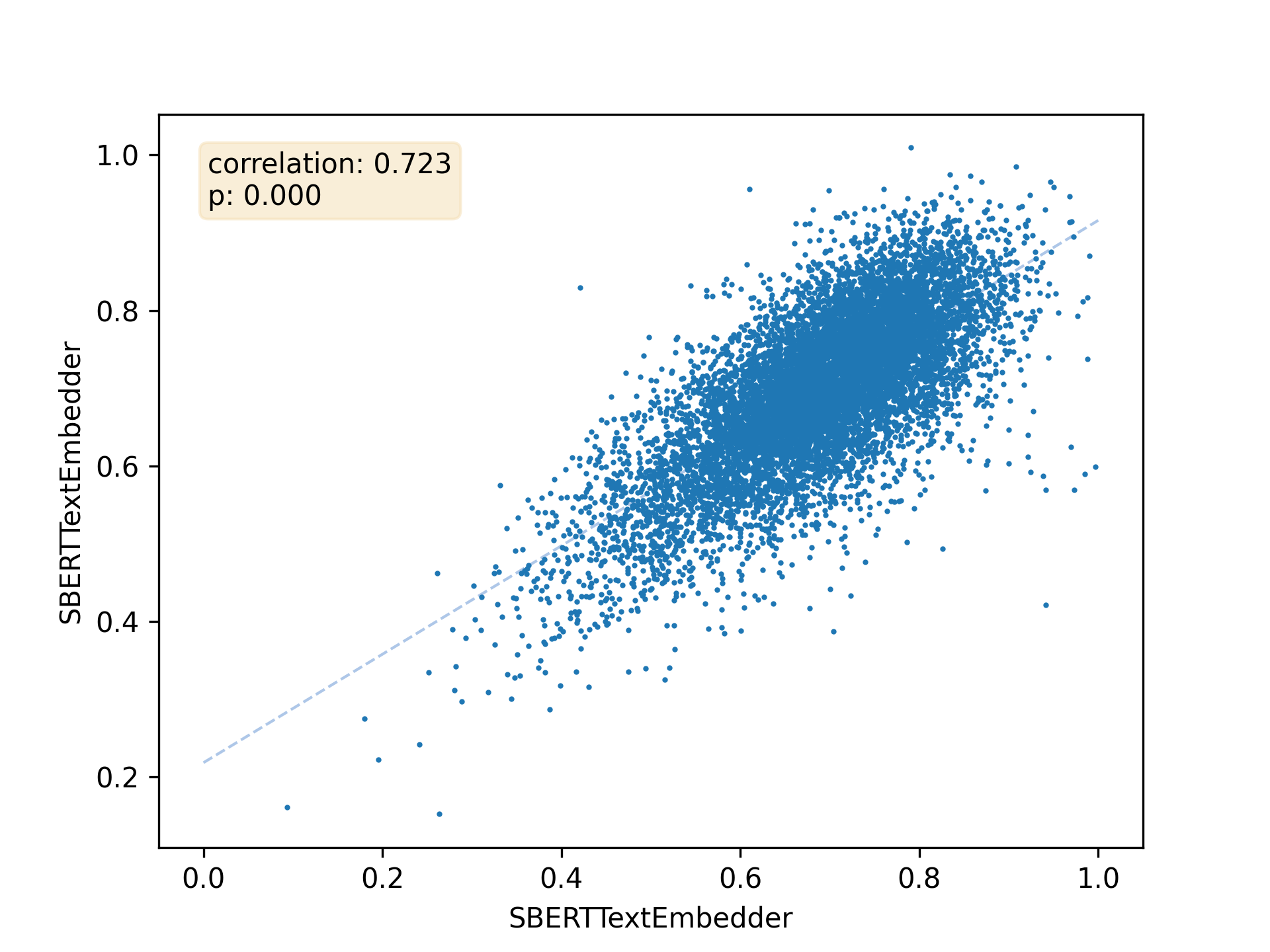}
    \caption{BillSum dataset results showing a strong, positive correlation.}
    \label{fig:billsum-results}
\end{subfigure}
\hfill
\begin{subfigure}{0.4\textwidth}
  \includegraphics[width=2.5in]{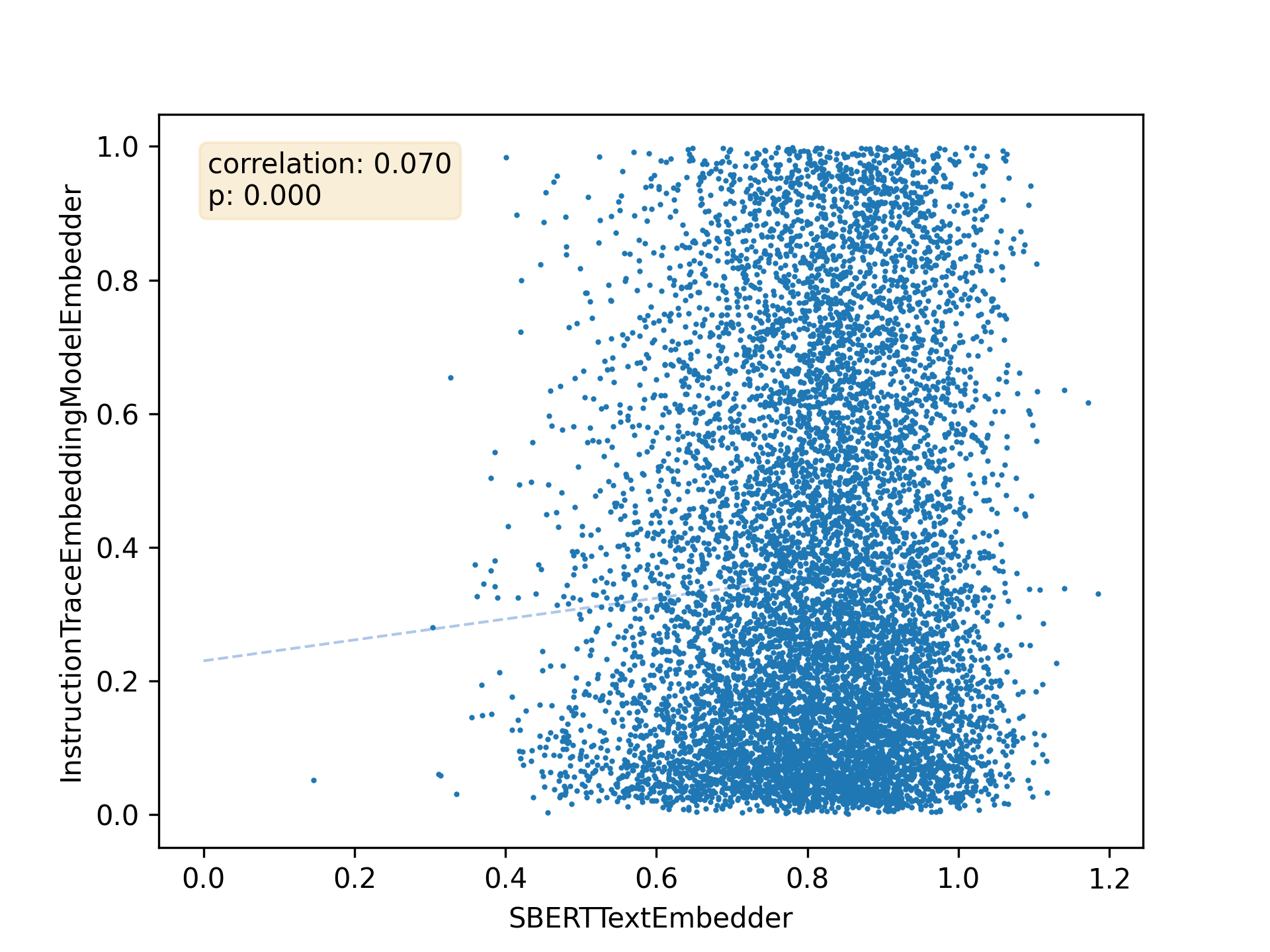}
  \caption{Stack Overflow dataset results showing virtually no correlation}
  \label{fig:stackoverflow-results}
\end{subfigure}
\hfill
\caption{EDC results for BillSum and Stack Overflow Datasets}
\label{fig:figures}
\end{figure}

Evaluating the quality of a dataset for training a machine learning model is a
task for which there is no standard methodology.  
To our knowledge, no previous work has used a methodology similar
to ours to evaluate sequence-to-sequence, summarization-style datasets.  In
order to validate our methodology and to understand what ``good'' performance
looks like on a high-quality dataset, we first performed an analysis of the
BillSum dataset \cite{billsum}.  BillSum is a dataset of over 20,000 U.S.
Congressional bills and reference summaries that is often used as a text
summarization benchmark and is widely cited by text summarization papers.
Using the SBERT text embedding model, we generated and compared
embeddings for the full text and the summary of a given bill in the dataset -
the results are depicted in Figure~\ref{fig:billsum-results} and clearly shows 
a strong correlation, which accords with the consensus that this is a good dataset.
We also conducted an experiment in which we randomly degraded a fixed percentage
of that dataset by reassigning their summary labels to another random sample. We
created several degraded datasets with different degradation percentages and
evaluated each using our EDC method. The results in
Table~\ref{table:DegredationExperiment} show that as the dataset quality goes
down, the EDC score decreases as expected.

\begin{table}[h!]
\centering
\caption{Methodology Validation: Degradation Experiment}
\label{table:DegredationExperiment}
\begin{tabular}{|c|c|c|} \hline
\textbf{Dataset} & \textbf{Degredation} & \textbf{EDC Score (correlation)} \\ \hline\hline
billsum & 0\% & 0.723 \\ \hline
billsum & 20\% & 0.487 \\ \hline
billsum & 40\% & 0.326 \\ \hline
billsum & 60\% & 0.180 \\ \hline
billsum & 80\% & 0.047 \\ \hline
billsum & 100\% & -0.014 \\ \hline
\end{tabular}
\end{table}

\subsubsection{Dataset Evaluation}
\label{sec:ourDatasetResults}

We evaluated three different datasets using EDC: HumanEval-X,
XLCost, and our own Stack Overflow dataset (see Sections 
\ref{sec:existing-datasets} and \ref{sec:constructed-dataset}). The 
first of these showed a weak, positive correlation (0.219) between the 
binary code and the text summary embedding distances, with a p-value 
less than 0.05. The correlation is much lower than our baseline BillSum
dataset. We thus deem this a low-quality dataset; the summarization 
problem it represents would be difficult for a model to learn. 
XLCost performs even worse, almost no correlation (0.066) 
between embedding distances, with a p-value less than 0.05. We judge 
this is a very low-quality dataset and also not likely to be learnable. 
Additionally, we observed ``binning'' for this dataset, where code distance 
scores tended to cluster around a few set points. This might suggest
issues with the diversity of the dataset (multiple input or output 
embeddings very close to one another) that should be investigated further.
Finally, performance on the Stack Overflow dataset created by the authors is depicted in
Figure~\ref{fig:stackoverflow-results}. Similar to XLCost, we observe little 
to no correlation between the code and text summary embedding distances. 
While this dataset does not have the binning problem observed with XLCost, 
we judge it to be low quality as the correlation is so low with respect to 
the baseline of BillSum, a known good dataset. Unfortunately, we were forced
to conclude that none of these datasets were viable choices to proceed with
training a transformer model.

\subsubsection{Human Expert Sanity Check}
\label{sec:SOSurvey}


\begin{table}
\centering
\caption{Survey Results}
\label{table:surveyResults}
\begin{tabular}{|c|c|c|c|} 
\hline
\textbf{Embedding} & \textbf{c(Agree)} & \textbf{c(Unsure)} & \textbf{c(Disagree)} \\ \hline\hline
binary &     42 &    12 &   46 \\ \hline
text &       20 &    10 &   70 \\ \hline
\end{tabular}
\end{table}

The results of the manual survey of the embedding distances for the Stack Overflow dataset 
appear in Table~\ref{table:surveyResults}.
Adjudicators agreed with the binary embedding distances only 42\% of the time,
and with the text embedding distances only 20\% of the time.  This seems strong
evidence for something being wrong either with the embedding or with the
dataset.  Anecdotally, adjudicators noted that they could sometimes see why the
text embedding might be judged similar, such as when both contained prefatory
remarks like ``This is for an algorithms class and I can't get my code to
compile.'' These aren't really statements about the probable business of the 
code yet they might make up a large fraction of the prose, thus diluting 
the dataset. 

\subsubsection{Off-the-shelf Solutions}
\label{sec:GPT3Results}

The release of ChatGPT~\footnote{https://openai.com/blog/chatgpt}, a
language model designed to interpret and answer questions in an conversational
fashion, has led researchers in the field of program analysis
to wonder if it can be used to summarize code as an aid to reverse engineering.
GPT-based solutions, if
sufficiently capable of summarizing even binary code without any additional training, would
seem to obviate the need for a dedicated binary code summarization model (and
consequently the need for a high-quality code summarization dataset).

In order to evaluate GPT-based binary code summarization, we used the dataset 
with highest EDC score: HumanEval-X.  To measure performance, we queried GPT-3 
with the disassembly of the binary samples along with a prompt asking for a
detailed summary of the code's functionality and informing GPT-3 that the
provided assembly was for the x86 architecture and in Intel syntax.  Next,
we computed pairwise distances between the GPT-3 code summaries and between 
the ground truth, human generated code behavior annotations in the HumanEval-X 
dataset, using the SBERT text embedding model again. We then computed the
correlation between these two sets of distances to measure the overall
performance of the GPT-3 code summaries, in a manner analogous to our EDC methodology.

There is a very weak correlation (0.056) between the similarity of the GPT-3
summaries and ground truth data, with a p-value less than 0.05. GPT-3 does a
very poor job of summarizing the code in this dataset. Upon closer inspection,
we identified a few failure classes exhibited by the GPT-3 model for code
summarization that are worth highlighting. 

The most common type of failure that we saw consisted of a coherent and
somewhat convincing description of a program that had nothing to do with the
sample provided.  This is a common problem when using large language models to
answer concrete questions and is often referred to as model ``hallucination''.
An example of this appears in Table~\ref{table-chat-gpt-fails} in the line labeled ``HALL''
This error case would be particularly bad if presented to a user that is
unfamiliar with reverse engineering who is unlikely to be able to verify the
code summary and might be easily convinced by a such a confident-sounding
summary.
The next most common type of failure that we saw consisted of an overly generic
and largely unhelpful description of the sample code.  For example, one sample
was summarized by GPT-3 as:
An example of this kind of error appears in the same table, in the line labeled ``GEN''.
This auto-generated description is largely unhelpful and even appears to contain some text
which was copied verbatim from the x86 instruction manual.
A third kind of failure also manifested itself, in a small number of cases, 
where the the model did not return readable text at all.
Instead, it  returned a subset of the provided assembly with no additional
context or summary.

\begin{table}[h!]
\centering
\caption {Sample GPT-3 Generated Binary Summaries}
\label{table-chat-gpt-fails}
\begin{tabular}{|l|l|l|} \hline
Label & GPT-3 Summary &                                  HumanEval-X Summary \\ \hline
HALL &  This is a function that calculates the           & Given a positive integer n, \\
& n-th Fibonacci number. It begins by saving the base    & return the count of the numbers \\
& pointer registry(RBP) onto the stack using the PUSH    & of n-digit positive integers \\
& instruction and assigns the stack pointer value to the & that start or end with 1... \\
& RBP registry with MOV... &  \\ \hline
GEN & This assembly code snippet is performing a calculation & Return median of elements in \\
& The calculation breaks down into 5 steps: 1. The           & the vector l. \\
& code begins with the instruction endbr64, which allows & \\
& for 64 bit indirect branch enforcement when executed &  \\
& in hardware. This instruction has been added in the  & \\
& x86 instruction set to Offer Enhanced Security...& \\  \hline
\end{tabular}
\end{table}

\section{Future Work and Recommendataions}\label{sec:FutureWork}
This is a work in progress.
We believe that the Embedding Distance Correlation (EDC) method for evaluating the \emph{quality} of a dataset is valuable and novel and are excited to present it.
We will be using it extensively in future efforts, both to evaluate datasets we assemble, as well as those made available by other researchers.
Our next steps in this project will be to devise procedures to assemble larger and higher quality datasets.
One approach would be to improve an existing dataset by data augmentation.
This might mean distilling English descriptions via summarization.
Or we might filter or weight ``good'' exemplars identified using our own EDC method.
Another possibility is amplifying the effect of known good exemplars by automatically re-writing their explanations or generating equivalent code versions via source transformation obfuscation.

We will make any datasets generated in this work available to researchers, including the Stack Overflow dataset. 
We will also be on the lookout for new datasets made available by other researchers and will assess their quality via EDC.
More detail is available in an extended Technical Report\cite{davis23}.

\bibliographystyle{splncs04}
\bibliography{sections/biblio}

\end{document}